# Handwritten Character Recognition Using Unique Feature Extraction Technique

Sai Abhishikth Ayyadevara, P N V Sai Ram Teja, Bharath K P Rajesh Kumar M *Senior Member, IEEE*
*School of Electronics Engineering*
*Vellore Institute of Technology*
Vellore, India

abhishikth58@gmail.com, peralisairam1997@gmail.com, bharathkp25@gmail.com, mrajeshkumar@vit.ac.in

*Abstract*—one of the most arduous and captivating domains under image processing is handwritten character recognition. In this paper we have proposed a feature extraction technique which is a combination of unique features of geometric, zone-based hybrid, gradient features extraction approaches and three different neural networks namely the Multilayer Perceptron network using Backpropagation algorithm (MLP BP), the Multilayer Perceptron network using Levenberg-Marquardt algorithm (MLP LM) and the Convolutional neural network (CNN) which have been implemented along with the Minimum Distance Classifier (MDC). The procedures lead to the conclusion that the proposed feature extraction algorithm is more accurate than its individual counterparts and also that Convolutional Neural Network is the most efficient neural network of the three in consideration.

## I. Introduction

Due to the recent developments in the image processing industry and the overall technological advancements the world has seen over the last couple of decades, handwritten character recognition has been gaining a lot of importance both in professional and personal lives. Though the existing feature extraction approaches are very effective, there is always a need to create better methods. Some existing techniques include geometric feature extraction, gradient feature extraction, zone based hybrid extraction and many more. Along with feature extraction another instrumental stage of the handwritten character recognition process is the usage of artificial neural networks. There are plenty of existing neural networks today, and all of them use various different algorithms to approach their job. Some of the existing neural networks include the Multilayer Perceptron, the Convolutional Neural Network and the likes. The recent literature describes these feature extraction and artificial neural networks. S. Kowsalya et al **[1]** presented the recognition of characters of the Tamil language. An Effective Learning Machine (ELM) was used to train the program for each character's geometric features. L. Anlo Safi et al **[2]** presented an overview of Feature Extraction techniques for offline recognition of Tamil characters. This paper proposes Zone-based hybrid feature extraction technique to achieve its goal. The features which were extracted from the character image were the number of horizontal, vertical, diagonal lines along with their total length for each zone. An Artificial Neural Network was used for classification and recognition purpose. Additionally Rajasekar M. et al **[3]** presented a paper that deals with the effects of changing the types of Neural Networks to achieve optical character recognition. They then analyze the results to determine which form of neural network technique is the most effective. The different parameters associated with each neural network are accounted to reach a conclusion. Dr. P Bhanumati et al **[4]** presented a paper in which feature extraction using gradient feature extraction technique was performed with the aid of an Artificial Neural Network. S.M Shiny et al **[5]** presented a paper that aims to extract the features of a text containing image using Sub-line Direction and Bounding Box algorithms. In order to maximize the efficiency of the recognition process, the Support Vector Machine has been implemented. Prashanth Vijayaraghavan and Misha Sra **[6]** presented a technique to classify characters using convolutional neural networks (ConvNets) into 35 different classes. They augment the ConvNetJS library for learning features by using stochastic pooling, probabilistic weighted pooling, and local contrast normalization. J. Sutha et al **[7]** presented the usage of structure analysis of Multilayer Perceptron network for handwritten Tamil character recognition using Levenberg-Marquardt Algorithm. A Lawgali et al **[8]** compared the effectiveness of Discrete Cosine Transform and Discrete Wavelet transform to capture discriminative features of Arabic handwritten characters.

In this paper a unique feature extraction technique, which originated from select characteristics of existing techniques is proposed. Also three different neural networks are compared and a conclusion is reached on which of these neural networks in the most efficient of the lot.

## II. METHODOLOGY

The following algorithms have been presented in this paper:

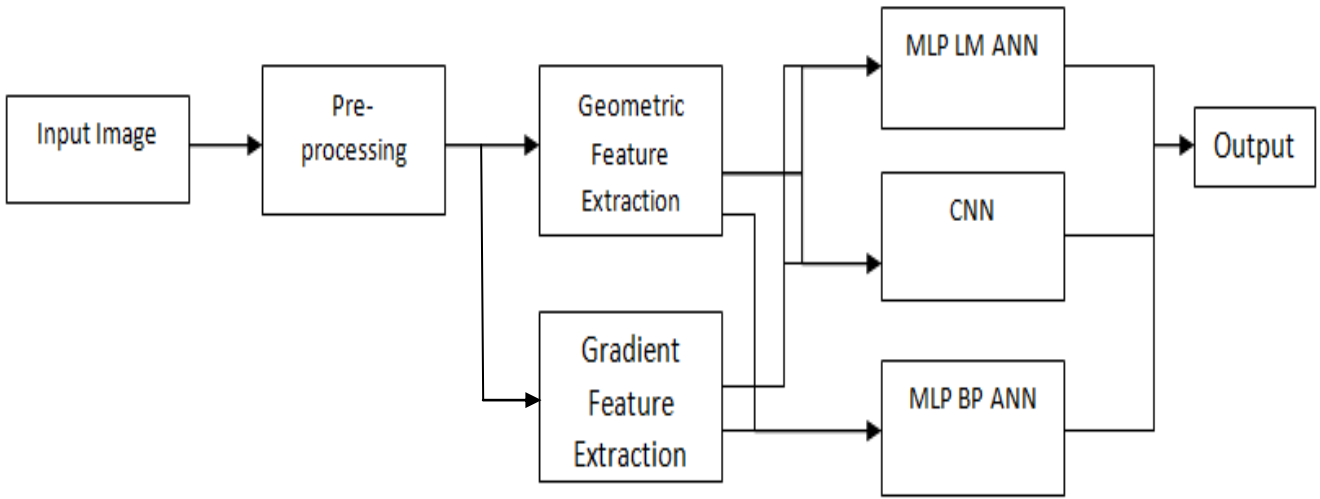

Figure 1: Neural Network Comparison Algorithm

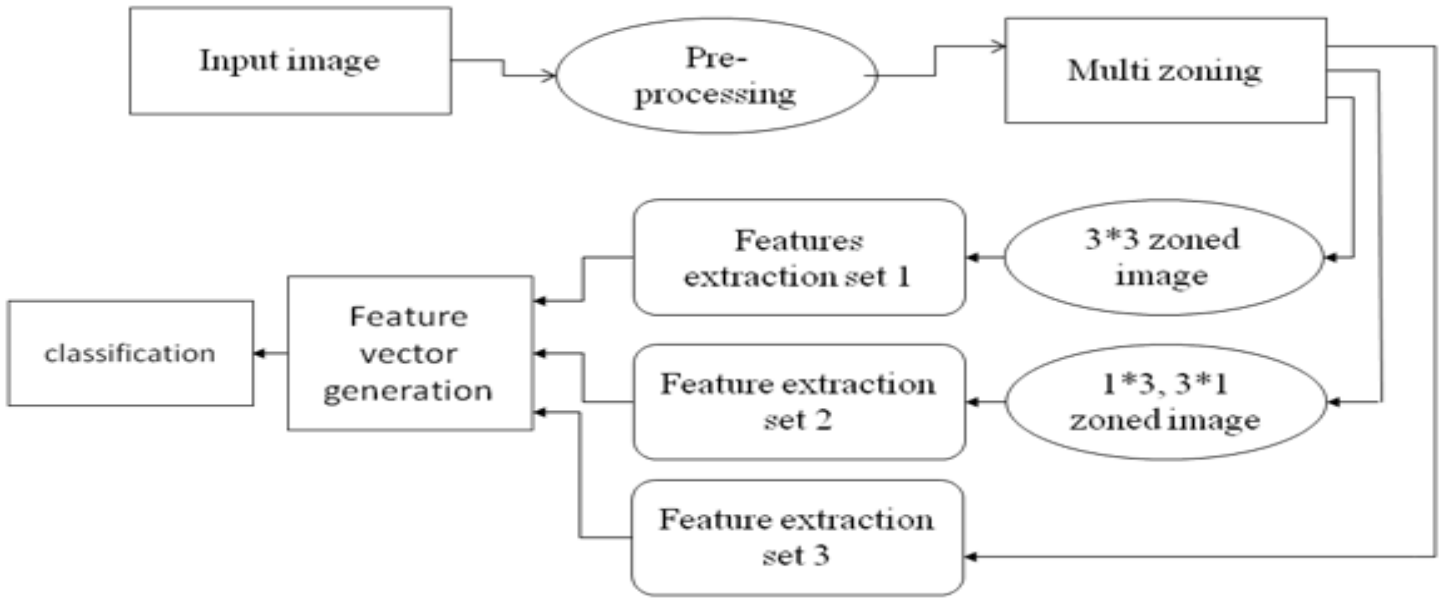

Figure 2: Proposed Extraction Technique's Algorithm

The above algorithms have been described in detail below.

### A. Preprocessing

Both the algorithms implemented the same preprocessing where the input image is first converted to grayscale image. Then the image is binarized followed by skeletonization. Then bounding box is applied on the image and is given as input for feature extraction. Binarisation is the process of differentiating the foreground from the background by selecting a threshold value which helps to ease the feature extraction process. The gray scale image is binarised by Otsu's global thresholding method. Otsu's method automatically predicts the threshold value which minimizes the weighted with-in class variance.

### B. Feature Extraction

The characteristics that form the proposed feature extraction technique have been illustrated as follows.
Multi-zoning: The pre-processed image obtained from the preprocessing block was zoned into sub-images with 3*3, 1*3,

3*1 configurations and features were extracted from each of these zones to achieve an improved accuracy.

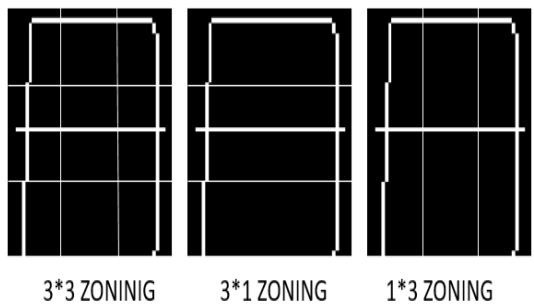

Figure 3: 3*3, 3*1, 1*3 zoning configurations

3 sets of features were extracted in feature vector generation. 3*3 configurations were the source of the first set of features. The second set of features was extracted from the zones of 1*3 and 3*1 configurations. The entire image played a role is the extraction of feature set 3.

*Set of features extracted per zone in feature set 1*:

Number of horizontal lines, number of vertical lines, number of right-diagonals, number of left-diagonals, the normalized lengths of these lines and the normalized skeleton area of the character.

*Set of features extracted per zone in feature set 2*:

Number of horizontal, vertical, right-diagonal, left-diagonal lines and normalized lengths of these line, normalized skeleton area of the character, number of intersections.

*Set of features extracted from image in feature set 3*:

Centroid, secondary-moments, number of objects, spread of character over back ground of the character image.

Equation (2) represents the number of lines and equation (3) represents the length of the character.

$$Value = 1 - 2 * \{\frac{Number\ of\ lines}{10}\} \qquad (2)$$

$$Value = \frac{Number\ of\ Pixels\ in\ Line}{Number\ of\ Pixels\ in\ the\ Corresponding\ Zone} \qquad (3)$$

Feature set 1 contains 81 different features as extracted equally from the 9 different zones of the image. 10 different features are extracted from three different zones to result in the 30 unique features that formed feature set 2. The pre-processed characters serve as an input to feature extraction set 3 resulting in 4 features. The concatenation of all the above features together sums up to 146 features. These 146 features form the feature vector of the corresponding character.

*C. Training and Testing the Artificial Neural Networks*

The artificial neural networks (ANN) were designed to duplicate the functions of a biological neural system. Generally an ANN consists of 3 different types of neurons:
- Input neurons
- Hidden neurons and
- Output neurons

The neural networks that are discussed in this paper are:
- Multilayer Perceptron Neural Network Using Backpropagation Algorithm
- Multilayer Perceptron Neural Network Using Levenberg-Marquardt Algorithm and
- Convolutional Neural Network

A multilayer perceptron is the fundamental kind of neural network. It consists of 3 layers or more (in this case 3) and except for the input node, all the other nodes are neurons that perform a nonlinear activation function.

a) The backpropagation algorithm

This is the most widely used algorithm and one of the easiest to implement. The backpropagation algorithm is used to figure out the value of a gradient that is in turn necessary for the estimation of the weights needed to be used in the multilayer perceptron neural network.

b) The Levenberg-Marquardt algorithm

Figure 3 provides an overview of the Levenberg- Marquardt algorithm

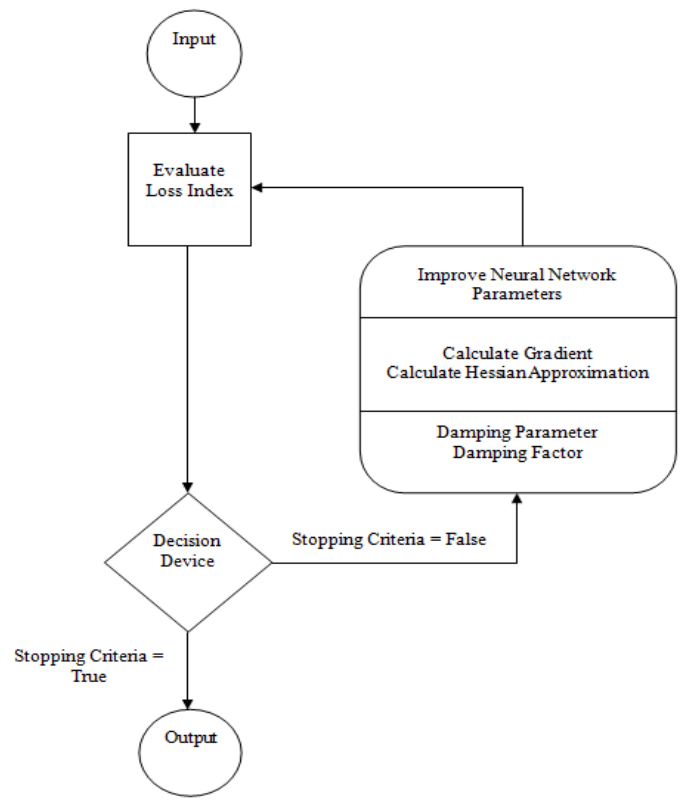

Figure 4: Levenberg- Marquardt Algorithm

c) Convolutional Neural Network

Also known as ConvNet and CNN, they are very similar to ordinary neural networks but differ in a small but crucial way. They are made up of neurons that are characterized by weights and biases both of which are trainable. An input is received into each neuron which then executes a dot product, sometimes succeeded by a non-linearity like any general neural network. The crucial difference is that CNNs make the blunt conjecture that the input is a picture which allows the user to encode some characteristics into the network. These make the forward function less error-prone and decrease the number of parameters in the network.

### III. RESULTS AND DISCUSSIONS

The proposed algorithm is trained with 78 images and tested on 26 images. With 78 character images the training matrix has dimensions of 145*78 and was used for classification of the test character. A performance analysis was done using the above stated training and testing sets on the proposed algorithm and the obtained accuracies were compared to that of geometric, zone-based hybrid and gradient features extraction methods.

From TABLE I. it can be inferred that the proposed algorithm shows a marked improvement in comparison to its parent algorithms.

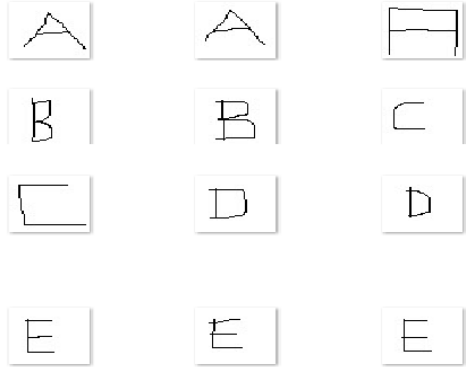

Figure 5: Training set

TABLE I. PERFORMANCE ANALYSIS OF DIFFERENT FEATURE EXTRACTION ALGORITHMS

| Features extraction method used | Classifier used | Accuracy |
|---|---|---|
| Gradient | MDC | 80.77% |
| Zone-based hybrid | MDC | 84.61% |
| Geometric | MDC | 80.77% |
| Proposed algorithm | MDC | 88.46% |

The algorithm has then been tested on select individual characters ('A', 'L' and 'Z') and a similar comparison has been drawn out as illustrated by TABLE II.

TABLE II. PERFORMANCE ANALYSIS OF DIFFERENT FEATURE EXTRACTION ALGORITHMS ON INDUVIDUAL CHARACTERS

| Features extraction | Classifier used | A | L | Z |
|---|---|---|---|---|
| Geometric | MDC | 90% | 80% | 100% |
| Zone-based hybrid | MDC | 80% | 100% | 90% |
| Gradient | MDC | 100% | 70% | 90% |
| Proposed algorithm | MDC | 100% | 70% | 100% |

TABLE III illustrates the performance analysis of the different feature extraction techniques when they were implemented on a dataset of different marker size than the one used before.

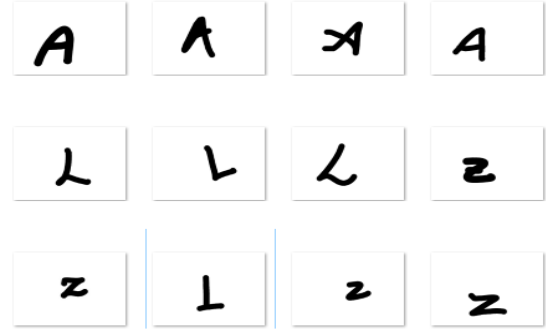

Figure 6: Testing Set with Different Marker Size

TABLE III. PERFORMANCE ANALYSIS OF DIFFERENT FEATURE EXTRACTION ALGORITHMS ON INDUVIDUAL CHARACTERS

| Features extraction | Classifier used | A | L | Z |
|---|---|---|---|---|
| Geometric | MDC | 70% | 60% | 60% |
| Zone-based hybrid | MDC | 30% | 40% | 100% |
| Gradient | MDC | 100% | 60% | 10% |
| Proposed algorithm | MDC | 90% | 70% | 70% |

Then the three neural networks were trained and tested using the dataset that was first used to test the feature extraction techniques. The accuracies for all three neural networks were recorded, first after implementing geometric feature extraction, and then by using gradient based feature extraction.

The results have been tabulated below.

TABLE IV. GEOMETRIC FEATURE EXTRACTION

| Neural Network: | No. of Training Characters | No. of Testing Characters | Accuracy |
|---|---|---|---|
| MLP BP | 104 | 52 | 86.5385% |
| MLP LM | 104 | 52 | 88.4615% |
| CNN | 104 | 52 | 88.4615% |

TABLE V. GRADIENT FEATURE EXTRACTION

| Neural Network: | No. of Training Characters | No. of Testing Characters | Accuracy |
|---|---|---|---|
| MLP BP | 104 | 52 | 84.6154% |
| MLP LM | 104 | 52 | 90.3846% |
| CNN | 104 | 52 | 92.3077% |

As one can observe from the results obtained, the feature extraction procedures do impact the final accuracy of recognition, but it is not possible to predict the manner in which they do. Although the Levenberg- Marquardt algorithm and the convolutional neural network achieved similar accuracies when geometric feature extraction was performed, the latter outdid the former when gradient feature extraction was considered.

## IV. CONCLUSION

In this paper, two different proposals were put forward. The first one was a new feature extraction technique which combined select characteristics of three different existing feature extraction techniques. The results showed that the proposed technique was comparatively more efficient. The demerits of the existing algorithms are that they consider a restricted number of features to aid their extraction process. The proposed algorithm on the other hand, uses the best of all these algorithm to create a more robust extraction technique trumping the existing ones in the process.

The second one was a performance analysis of three different neural networks for two different feature extraction techniques – geometric and gradient. The neural networks were: the MLP neural network using backpropagation algorithm, the MLP neural network using Levenberg-Marquardt algorithm and the CNN. The results portrayed that the CNN was the most efficient followed closely by the Levenberg-Marquardt algorithm.

## REFERENCES

[1] S. Kowsalya and P.S. Periyasamy "Handwritten Tamil Character Recognition Using Geometric Feature Extraction Approach" Asian Journal of Information Technology (AJIT) 15 (20); 4124-4128, 2016

[2] L. Anlo Safi and K.G. Srinivasagan "Offline Tamil Handwritten Character Recognition using Zone based Hybrid Feature Extraction Technique" International Journal of Computer Applications (IJCA) (0975 – 8887) Volume 65– No.1, 2013

[3] Rajasekar M.,Celine Kavida A and Anto Bennet M " Performance and Analysis of Handwritten Tamil Character Recognition using Artificial Neural Networks" International Journal of Recent Scientific Research Vol. 7, Issue,1, pp. 8611-8615, 2016

[4] Dr. P. Bhanumati, Mr. M. Ashraf, Mr. K. Gokul Kumar and Mr. G. Ragul "Tamil Handwritten Character Recognition Based On Gradient Features" International Journal of Recent Innovation in Engineering and Research (IJRIER), Scientific Journal Impact Factor - 3.605 by SJIF, e-ISSN: 2456 – 2084, 2017

[5] S.M. Shiny, M. Anthony Robert Raj and S. Abirami "Offline Tamil Handwritten Character Recognition Using Sub Line Direction and Bounding Box Techniques" Indian Journal of Science and Technology (IJST), Vol 8(S7), 110–116, 2015

[6] Prashanth Vijayaraghavan, Misha Sra "Handwritten Tamil Recognition using a Convolutional Neural Network" MIT MediaLab unpublished

[7] J. Sutha, N. Ramaraj "Structure Analysis of Multilayer Perceptron Network for Handwritten Tamil Character Recognition Using Levenberg-Marquardt Algorithm" International Journal of Soft Computing (IJSC) 3 (5); 373-381, 2008

[8] A. Lawgali, A. Bouridane, M. Angelova, Z. Ghassemlooy "Handwritten Arabic Character Recognition: Which Feature Extraction Method?" International Journal of Advanced Science and Technology (IJAST) Vol. 34, 2011